\pdfoutput=1

\documentclass[11pt]{article}

\usepackage{acl}

\usepackage{times}
\usepackage{latexsym}

\usepackage[T1]{fontenc}

\usepackage[utf8]{inputenc}

\usepackage{microtype}

\usepackage{soul}
\usepackage{url}

\usepackage{amssymb}

\usepackage{graphicx}
\usepackage{amsmath}
\usepackage{amsthm}
\usepackage{booktabs}
\usepackage{paralist}
\usepackage{color}

\usepackage{algorithmic}
\usepackage{algorithm2e}
\newtheorem{myDef}{Definition}

\usepackage{bbding}
\usepackage{multirow}

\newcommand{\secref}[1]{Section~\ref{#1}\xspace}
\newcommand{\exam}[1]{\emph{#1}\xspace}
\newcommand{\vect}[1]{\mathbf{#1}\xspace}
\newcommand{\vecti}{\vect{i}\xspace}
\newcommand{\vectj}{\vect{j}\xspace}
\newcommand{\vectk}{\vect{k}\xspace}
\newcommand{\vects}{\vect{s}\xspace}
\newcommand{\vectr}{\vect{r}\xspace}
\newcommand{\vecto}{\vect{o}\xspace}

\definecolor{myblue}{rgb}{0.13, 0.67, 0.8}
\definecolor{mygreen}{rgb}{0.55, 0.71, 0.0}
\definecolor{myyellow}{rgb}{1.0, 0.49, 0.0}
\definecolor{myred}{rgb}{0.82, 0.1, 0.26}

\newtheorem{theorem}{Theorem}
\newtheorem{lemma}{Lemma}
\usepackage{tabu}
\usepackage{microtype}
\usepackage{subfigure}

\title{RotateQVS: Representing Temporal Information as Rotations in Quaternion Vector Space for Temporal Knowledge Graph Completion}

\author{
Kai Chen$^1$,
Ye Wang$^{1,2}$\thanks{\quad Corresponding author.} \ ,
Yitong Li$^3$,
Aiping Li$^1$
\\
$^1$National University of Defense Technology, Changsha, China\\
$^2$Pengcheng Laboratory, Shenzhen, China\\
$^3$Huawei Technologies Co., Ltd.\\
{\small \texttt{\{chenkai\_,ye.wang,liaiping\}@nudt.edu.cn} } \quad {\small \texttt{liyitong3@huawei.com}}
}

\begin{document}
\maketitle
\begin{abstract}
Temporal factors are tied to the growth of facts in realistic applications, such as the progress of diseases and the development of political situation, therefore, research on Temporal Knowledge Graph (TKG) attracks much attention.
In TKG, relation patterns inherent with temporality are required to be studied for representation learning and reasoning across temporal facts.
However, existing methods can hardly model temporal relation patterns, nor can capture the intrinsic connections between relations when evolving over time, lacking of interpretability.
In this paper, we propose a novel temporal modeling method which represents temporal entities as \textbf{Rotations} in \textbf{Q}uaternion \textbf{V}ector \textbf{S}pace (RotateQVS) and relations as complex vectors in Hamilton’s quaternion space.
We demonstrate our method can model key patterns of relations in TKG, such as symmetry, asymmetry, inverse, and can further capture time-evolved relations by theory.
Empirically, we show that our method can boost the performance of link prediction tasks over four temporal knowledge graph benchmarks.
\end{abstract}

\section{Introduction}

Knowledge Graphs (KGs) have been widely adopted to represent informative knowledge or facts in real-world applications \cite{Bollacker2008FreebaseAC,miller1995wordnet,suchanek2007yago}.
However, as known facts are usually sparse, KGs are far from completeness. Thus, Knowledge Graph Completion (KGC) methods are proposed to predict missing facts, i.e. links between entities \cite{bordes2013translating,Yang2015EmbeddingEA,Dettmers2018Convolutional,CIM}.
Furthermore, in real world, many facts are bonded with a particular time by nature.
For example, \exam{Barack Obama is the president of USA} is only valid for the time period \exam{2009 - 2017}.
To model such time-sensitive facts, Temporal Knowledge Graphs (TKGs) have recently drawn growing attention from both academic and industrial communities \cite{lautenschlager2015icews,leetaru2013gdelt}.

TKG Embedding (TKGE) methods \cite{jiang2016towards,dasgupta2018hyte,jin2020recurrent,ChronoR} were proposed to represent entities and relations with temporal features in TKGs  \cite{lautenschlager2015icews,leetaru2013gdelt}.
But how to present them with temporal interpretability remains a challenge for state-of-the-art TKGE models.
Further, it is crucial for TKG Completion (TKGC) to leverage the learned temporal information.
Previous static KGC works \cite{sun2019re,Schlichtkrull2018Modeling,gao2020rotate3d} learn explainable embeddings of various relation patterns, so that symmetric pattern (e.g. ``co-author''), asymmetric pattern (e.g. ``affiliation''), inverse pattern (e.g. ``buyer'' vs. ``seller'') and complex composition pattern (e.g. ``father's wife (mother)'' vs. ``wife's father (father in law)'') can be captured in static KGs. However, in TKGs, there are inherent connections between entities and their relations along with time-evolving.
For example, the relation between \textit{Kit Harington} and \textit{Rose Leslie} is \textit{in love} in \textit{2012}, becomes \textit{engaged} in \textit{2017}, and then turns into \textit{married} in \textit{2018}.
To the best of our knowledge, very few of the existing TKGE methods can capture them.

To address this problem, we take inspirations from Hamilton's quaternion number system \cite{Hamilton1844,zhang2019quaternion,gao2020rotate3d} and propose
a novel method based on quaternion. To be specific, we encode both entities and relations as quaternion embeddings, and then temporal entity embeddings can be represented as \textbf{Rotations} in \textbf{Q}uaternion \textbf{V}ector \textbf{S}pace (RotateQVS).
Theoretically, we show the limitations of previous methods and demonstrate that performing quaternion embeddings can model symmetric, asymmetric, and inverse relation patterns.
Meanwhile, we prove our method is capable of capturing time-evolving information in TKG explicably.
We empirically evaluate our method over four TKGC benchmarks and report state-of-the-art performance consistently.
Further, we perform analysis of the learned quaternion embeddings and show the abilities of our RotateQVS for modeling various relation patterns, including temporal evolution.

We summarize our main contributions as follows:
\begin{enumerate}
\item We propose an original quaternion based TKGC method, namely RotateQVS, which represents temporal information as rotations in quaternion vector space.

\item We study temporal evolving relations, and we demonstrate the proposed RotateQVS can model various relation patterns including temporal evolution both theoretically and empirically.

\item Our RotateQVS outperforms the SOTA methods over all of ICEWS14, ICEWS05-15, YAGO11k and GDELT datasets on link prediction task.
\end{enumerate}

\section{Preliminaries on Hamilton’s Quaternions}
\label{Quaternion}

Quaternion number system \cite{Hamilton1844} is an extension of traditional complex numbers.
Recently, quaternion has been applied in static knowledge graph embedding~\cite{zhang2019quaternion,gao2020rotate3d}.
For readers better understanding our method in \secref{sec:our_method}, we introduce the definition and basic operations of quaternion in this section.

\subsection{Quaternion Operations}
A quaternion is expressed as $q = a + b \vecti +c \vectj +d \vectk$, and some key quaternion operations are defined as:

\paragraph{Conjugate}
Similar to a traditional complex number, the conjugate of a quaternion is defined with the same real part and the opposite imaginary parts, that is $$\overline{q} = a - b \vecti -c \vectj -d \vectk \, . $$

\paragraph{Inner Product}
The inner product between $q_1 = a_1 + b_1 \vecti + c_1 \vectj + d_1 \vectk$ and $q_2 = a_2 + b_2 \vecti + c_2 \vectj + d_2 \vectk$ is the sum of product of each corresponding factor
\begin{equation} \nonumber
q_1 \cdot q_2 =	\langle a_1, a_2 \rangle + \langle b_1, b_2 \rangle + \langle c_1, c_2 \rangle + \langle d_1, d_2 \rangle \, .
\end{equation}

\paragraph{Norm}
With the definition of conjugate and inner product, the norm of a quaternion is defined as:
\begin{equation}\label{equation:norm}
    \vert \vert q \vert \vert = \sqrt{q \cdot \overline{q}} = \sqrt{\overline{q} \cdot q} = \sqrt{a^{2} + b^{2} + c^{2} +d^{2}}
\end{equation}

\paragraph{Inverse}
The inverse of a quaternion is defined from $q^{-1} \cdot q = q \cdot q^{-1} = 1$.
Multiplying by $\overline{q}$, we have $\overline{q} \cdot q \cdot q^{-1} = \overline{q}$, derived from which we get:
\begin{equation}\label{equation:inverse_q}
    q^{-1} = \frac{\overline{q}}{{\vert \vert q \vert \vert}^{2}}
\end{equation}

\paragraph{Hamilton Product}
For two quaternions $q_1$ and $q_2$, their product is determined by the products of the basis elements and the distributive law.
The quaternion multiplication formula is:
\begin{align}\label{equation:product}
    q_1  q_2 
    &= (a_1a_2 - b_1b_2 - c_1c_2 -d_1d_2) \nonumber\\
    &\quad + (a_1b_2 + b_1a_2 + c_1d_2 - d_1c_2)\vecti \nonumber\\
    &\quad + (a_1c_2 - b_1d_2 + c_1a_2 + d_1b_2)\vectj \nonumber\\ 
    &\quad + (a_1d_2 + b_1c_2 - c_1b_2 + d_1a_2)\vectk
\end{align}

Considering the conjugate of Hamilton product, we can further deduce:
\begin{align}
    \overline{q_1  q_2} 
    & = \overline{q_2} \hphantom{.} \overline{q_1} \, , \nonumber \\
    \overline{q_1  q_2  q_3} 
    & = \overline{q_3} \hphantom{.} \overline{q_2} \hphantom{.} \overline{q_1} \, .
\end{align}

\subsection{3D Vector Space}
In fact, the imaginary part $b \vecti + c \vectj + d \vectk$ of a quaternion behaves like a vector $\vect{v} = (b, c, d)$ in a 3D vector space.
Thus, conveniently, we rewrite a quaternion using \textbf{imaginary vector}s:
\begin{equation}\label{vector}
    q =	a +  b \vecti + c \vectj + d \vectk = a + \vect{v} = (a, \vect{0}) + (0, \vect{v}) \, .
\end{equation}

\paragraph{Multiplication rule} The multiplication of two imaginary vectors $\vect{v}_1$ and $\vect{v}_2$ is
\begin{equation}\label{equation:vector_multiplication}
    \vect{v}_1 \vect{v}_2 = \vect{v}_1 \times \vect{v}_2 - \vect{v}_1 \cdot \vect{v}_2 \, ,
\end{equation}
where $\vect{v}_1 \times \vect{v}_2$ is vector cross product, resulting in a vector, and $\vect{v}_1 \cdot \vect{v}_2$ is the dot product, which gives a scalar.
Obviously, the multiplication of two imaginary vectors is non-commutative, as the cross product is non-commutative.

Thus, the multiplication of two quaternions can be rewritten in 3D vector perspective:
\begin{align}\label{equation:multiplication_3D}
    & q_1 q_2 = (a_1, \vect{v}_1)\hphantom{.} (a_2, \vect{v}_2) \nonumber\\
    = & (a_1a_2 - \vect{v}_1 \cdot \vect{v}_2, a_1\vect{v}_2 + a_2\vect{v}_1 + \vect{v}_1 \times \vect{v}_2 ) 
\end{align}

\section{Proposed Method}
\label{sec:our_method}
In this section, we introduce a novel temporal modeling approach for TKG by representing temporal information as \textbf{Rotations} in \textbf{Q}uaternion \textbf{V}ector \textbf{S}pace (RotateQVS).

\subsection{Notations}
Suppose that we have a temporal knowledge graph, noted as $\mathcal{G}$. We use $\mathcal{E}$ to denote the set of entities, $\mathcal{R}$ to denote the set of relations, and $\mathcal{T}$ to denote the set of time stamps.
Then, the temporal knowledge graph $\mathcal{G}$ can be defined as a collection of quadruples, noted as $(s, r, o, t)$, where a relation $r \in \mathcal{R}$ holds between a head entity $s\in \mathcal{E}$ and an tail entity $o\in \mathcal{E}$ at time $t$.
The actual time $t$ is represented by a time stamp $\tau \in \mathcal{T}$.

\subsection{Representing Temporal Information using Rotations in 3D Vector Space}
Similar to Tero \cite{xu2020tero} which utilizes a rotation in complex space, we also represent temporal information using rotations while in the quaternion vector space.

In 3D vector space, according to \emph{Euler's rotation theorem} \cite{euler1776novi,verhoeff2014euler}, any rotation or sequence of rotations of a rigid body or a coordinate system about a fixed point is equivalent to a single rotation by a given angle $\theta$ about a fixed axis (called the Euler axis) that runs through the fixed point.
And an extension of Euler's formula for quaternion can be expressed as follows:
\begin{equation}
\begin{aligned}
    q &= e^{\frac{\theta}{2} (v_x \vecti + v_y \vectj + u_z \vectk)} \\
    &= \cos \frac{\theta}{2} + (v_x \vecti + v_y \vectj + u_z \vectk) \sin \frac{\theta}{2} \, ,
\end{aligned}
\end{equation}
where $\vecti$, $\vectj$, $\vectk$ are unit vectors representing the three Cartesian axes.

\subsubsection{Representing Time, Entities, and Relations:}
Quaternions provide us with a simple way to encode this axis–angle representation in four numbers, and can be used to perform the rotation procedure in 3D vector space.
By doing so, we constrain the \textbf{time stamp} embedding $\pmb{\tau}$ as a unit quaternion as 
\begin{equation}\label{equation:tau}
    \pmb{\tau} = \cos \frac{{\theta}_{\tau}}{2} + \vect{u_{\tau}} \sin \frac{{\theta}_{\tau}}{2} \, ,
\end{equation}
where $\vect{u_{\tau}}$ is a unit vector in the quaternion space.
And for other elements of a quadruple $(s, r, o, t)$, based on the Hamilton's quaternions in Section \ref{Quaternion}, we map each of them to its base, which is a time-independent quaternion embedding:
\begin{align}
\vects & = \{ \vect{a}_s + \vect{b}_s \vecti + \vect{c}_s \vectj + \vect{d}_s \vectk \} \nonumber\\
\vectr & = \{ \vect{a}_r + \vect{b}_r \vecti + \vect{c}_r \vectj + \vect{d}_r \vectk \} \nonumber\\
\vecto & = \{ \vect{a}_o + \vect{b}_o \vecti + \vect{c}_o \vectj + \vect{d}_o \vectk \} \, ,
\end{align}
where $\vect{a}_{\{.\}},\vect{b}_{\{.\}},\vect{c}_{\{.\}},\vect{d}_{\{.\}} \in \mathbb{R}^{k}$.

\subsubsection{Temporal Entities:}
\begin{figure}[t]
\centering
\includegraphics[width=0.35\columnwidth]{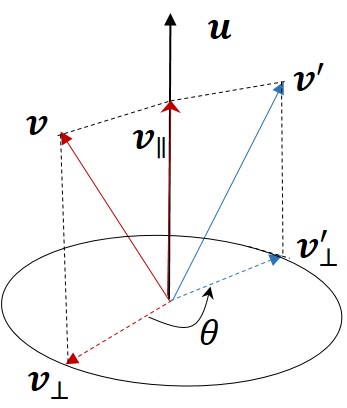} 
\caption{An illustration of the proposed rotation in 3D vector space, where $\vect{v'}$ is the result of vector $\vect{v}$ rotating $\theta$ around the rotation axis $\vect{u}$.}
\label{figure:rotation}
\end{figure}

We make use of the quaternion rules to represent temporal information as rotations in 3D vector space.
An abstract rotation procedure is illustrated in Figure~\ref{figure:rotation}.

\begin{theorem} \label{Theorem_rotation}
Given a unit quaternion $q = \cos \frac{\theta}{2} + \vect{u} \sin \frac{\theta}{2}$, where $\vect{u} \in  \mathbb{R}\vecti +  \mathbb{R}\vectj +  \mathbb{R}\vectk$ is a unit  vector (rotation axis) in a three-dimensional space, the result of vector $\vect{v}$ rotating $\theta$ around the rotation axis $\vect{u}$ is
\begin{equation}\label{equation:rotation}
    \vect{v'} = q	\vect{v} q^{-1} = q	\vect{v} \overline{q} \, .
\end{equation}
\end{theorem}

Theorem~\ref{Theorem_rotation} is supported by Rodrigues' rotation formula \cite{rodrigues1840lois}.\footnote{See proof in Appendix \ref{app:proof_theorem1}}
We then define the functional mapping that reflects the temporal evolution of an entity embedding. For each time stamp $\tau$, the functional mapping is an element-wise rotation from the basic entity embedding $\vect{e}$ (quaternion representation) to the \textbf{time-specific entity} embedding $\vect{e}_{t}$, which is as follows:
\begin{align}\label{equation:embedding}
    \vect{e}_{t} &= \pmb{\tau} \vect{e} \pmb{\tau}^{-1}
    = \pmb{\tau} (a_{\vect{e}} + {\vect{v_e}}) \pmb{\tau}^{-1} \nonumber\\
    &= a_{\vect{e}} \pmb{\tau} \pmb{\tau}^{-1} +  \pmb{\tau} {\vect{v_e}} \pmb{\tau}^{-1} \nonumber\\
    &= a_{\vect{e}} + \pmb{\tau} {\vect{v_e}} \pmb{\tau}^{-1} \, ,
\end{align}
where $a_{\vect{e}}$ and ${\vect{v_e}}$ are the scalar/real and vector/imaginary part of the entity quaternion representation $\vect{e}$ respectively. And according to Theorem \ref{Theorem_rotation}, $\pmb{\tau} {\vect{v_e}} \pmb{\tau}^{-1}$ is the result of vector $\vect{v_e}$ rotating ${\theta}_{\tau}$ around the rotation axis $\vect{u_{\tau}}$ ($\pmb{\tau} = \cos \frac{{\theta}_{\tau}}{2} + \vect{u_{\tau}} \sin \frac{{\theta}_{\tau}}{2}$, see Equation \ref{equation:tau}) which constitutes the vector/imaginary part of $\vect{e}_{t}$.
Thus, we can get a lemma:
\begin{lemma}\label{lemma:mapping}
The vector (imaginary) part is rotated while the scalar (real) part remains unchanged in the functional mapping (Equation~\ref{equation:embedding}) which reflects the temporal evolution of an entity embedding.
\end{lemma}

For a quadruple $(s, r, o, t)$, we make use of the functional mapping to get the time-specific entity embeddings $\vects_{t}$ and $\vecto_{t}$ from the basic entity embeddings $\vects$ and $\vecto$:
\begin{equation}
    \vects_{t} = \pmb{\tau} \vects \pmb{\tau}^{-1}, \quad
    \vecto_{t} = \pmb{\tau} \vecto \pmb{\tau}^{-1} \, .
\end{equation}

Considering the temporal evolution of entity embedding, the \textbf{relation} embedding $\vectr$ is regarded as a translation from the time-specific subject embedding $\vects_{t}$ to the conjugate of the time-specific object embedding $\vect{\overline{o_{t}}}$.
In other words, we aim to make $\vects_{t} + \vectr = \vect{\overline{o_{t}}}$ for all positive quadruples. Then, the score function can be defined as:
\begin{equation}\label{equation:score}
    f(s, r, o, t) =  \vert\vert\vects_{t} + \vectr - \overline{\vect{o}_t} \vert \vert \, .
\end{equation}
Note that each embedding above is a quaternion representation, and ``$\vert \vert$'' denotes the norm computation (see Equation~\ref{equation:norm}).
 
\subsubsection{Loss Function}
We use the same margin loss function with multiple negative sampling as proposed in \cite{SunDNT19}, which has been proved to be effective on distance-based KGE models \cite{bordes2013translating,SunDNT19} and as well as the TKGE models \cite{xu2019temporal,xu2020tero}.
In details, our loss function is
\begin{equation}
    \mathcal{L} = - \log\sigma (\gamma - f(\xi)) - \sum_{i=1}^{\eta}\log\sigma(f({\xi}'_i) - \gamma) \, ,
\end{equation}
where $\eta$ is the number of negative training samples over the positive one, $\xi$ is the positive training quadruple, $\sigma(\cdot)$ denotes the $\operatorname{sigmoid}$ function, $\gamma$ is a fixed margin, and $\xi'_i$ denotes the $i$-th negative sample generated by randomly corrupting the subject or the object of $\xi$ such as $(s', r, o, t)$ and $(s, r, o', t)$.

\subsection{Modeling Various Relation Patterns}

In this section, we demonstrate that our RotateQVS can model various relation patterns.
In TKGE, four kinds of relation patterns are mostly considered and studied in previous static KGE and TKGE works \cite{SunDNT19,gao2020rotate3d}.
Their definitions are given as follows:
\begin{myDef}
	\label{definition:r1}
    A relation $r$ is symmetric, if $ \forall s,o,t$, $ r(s,o,t) \land r(o,s,t)$  holds True.
\end{myDef}

\begin{myDef}
	\label{definition:r2}
    A relation $r$ is asymmetric, if $ \forall s,o,t$, $ r(s,o,t) \land \lnot r(o,s,t)$  holds True.
\end{myDef}

\begin{myDef}
	\label{definition:r3}
    Relation $r_1$ is the inverse of $r_2$, if $ \forall s,o,t$, $ r_1(s,o,t)$ $\land$$ r_2(o,s,t)$  holds True.
\end{myDef}

\begin{myDef}
	\label{definition:r4}
    Relation $r_1$ and $r_2$ are evolving over time from $t_1$ (time stamp $\tau_1$) to $t_2$ (time stamp $\tau_2$), if $ \forall s,o$, $ r_1(s,o,t_1) \land r_2(s,o,t_2)$  holds True.
\end{myDef}

Comparing with other TKGE methods, we show RotateQVS can model all these four patterns, while previous methods (see \secref{sec:Baselines}) fail to do so.\footnote{
Statistics of several baselines modeling on various relation patterns are summarised in Appendix \ref{app:Statistics}.}
One advantage of applying quaternion embeddings is that our method supports all these relation patterns, while other representation forms cannot, such as TeRo \cite{xu2020tero} using complex number system $a+b\vecti$.\footnote{Theoretical analysis of TeRo's defect is shown in Section~\ref{app:defect_TeRo}.}

As seen in our score function (Equation \ref{equation:score}), our aim is to make 
\begin{equation} \label{equation:relation_basic}
\begin{aligned}
& \pmb{\tau} \vects \pmb{\tau}^{-1} + \vectr = \overline{ \pmb{\tau} \vecto \pmb{\tau}^{-1}} = \pmb{\tau} \overline{\vecto} \pmb{\tau}^{-1}\\
& \Leftrightarrow 
  \overline{\vecto} - \vects  = \pmb{\tau}^{-1} \vectr \pmb{\tau} \, .
\end{aligned}
\end{equation}
Then we can get following results:

\begin{lemma}\label{lemma:symmetric}
RotateQVS can model the symmetric pattern for TKG. (See proof in Appendix \ref{app:proof_symmetric})
\end{lemma}

\begin{lemma}\label{lemma:asymmetric}
RotateQVS can model the asymmetric pattern for TKG. (See proof in Appendix \ref{app:proof_asymmetric})
\end{lemma}

\begin{lemma}\label{lemma:inversion}
RotateQVS can model the inversion pattern for TKG. (See proof in Appendix \ref{app:proof_inversion})
\end{lemma}

\begin{lemma}\label{lemma:temporal-evolution}
RotateQVS can model the temporal-evolution pattern for TKG.
\end{lemma}

\begin{proof}

For temporal-evolution pattern, $ r_1(s,o,t_1)$ $\land$ $ r_2(s,o,t_2)$ in Definition \ref{definition:r4} can be expressed as:
\begin{equation}\label{equation:te_proof}
\begin{aligned}
& \left\{  
     \begin{aligned}
     \overline{\vecto} - \vects  = \pmb{\tau}_1^{-1} \vect{r}_1 \pmb{\tau}_1 \\
    \overline{\vecto} - \vects  = \pmb{\tau}_2^{-1} \vect{r}_2 \pmb{\tau}_2
    \end{aligned}
\right. 
\\
& \Leftrightarrow
\pmb{\tau}_2 \pmb{\tau}_1^{-1} \vect{r}_1 (\pmb{\tau}_2 \pmb{\tau}_1^{-1})^{-1} = \vect{r}_2 \, .
\end{aligned}
\end{equation}
For the same head entity and tail entity, if a relation $r_1$ holds at time $t_1$ (time stamp $\tau_1$) and a relation $r_2$ holds at time $t_2$ (time stamp $\tau_2$), we are supposed to get  $\pmb{\tau}_2 \pmb{\tau}_1^{-1} \vect{r}_1 (\pmb{\tau}_2 \pmb{\tau}_1^{-1})^{-1} = \vect{r}_2$. 

In addition, based on Equation \ref{equation:te_proof}, we have
\begin{equation}\label{equation:temporal_form}
\pmb{\tau}_1^{-1} \vect{r}_1 \pmb{\tau}_1 = \pmb{\tau}_2^{-1} \vect{r}_2 \pmb{\tau}_2 \, .
\end{equation}
Since we have Theorem \ref{Theorem_rotation}, $\pmb{\tau}_1^{-1} \vect{r}_1 \pmb{\tau}_1$ and $\pmb{\tau}_2^{-1} \vect{r}_2 \pmb{\tau}_2$ can be regarded as rotations in quaternion vector space for $\vect{r}_1$ and $\vect{r}_2$, respectively, which indicates the norm of $\vect{r}_1$ is the same as that of $\vect{r}_2$. Furthermore, Lemma \ref{lemma:mapping} indicates the rotation mapping keeps the scalar/real part unchanged for a vector. Thus, we can have the following deductions:
\begin{equation}
\label{equation:further_deductions}
\begin{aligned}
     \left\{  
     \begin{aligned}
    & \vert \vert \vect{r}_1 \vert \vert = \vert \vert \vect{r}_2 \vert \vert \\
   & Re(\vect{r}_1) =  Re(\vect{r}_2) \,.
    \end{aligned}
\right. 
\end{aligned}
\end{equation}

Notice that Equation \ref{equation:further_deductions} is a sufficient and unnecessary conclusion of Equation \ref{equation:temporal_form}.
\end{proof}

\begin{table}[!t]
\centering
\resizebox{0.7\columnwidth}{!}{
\begin{tabular}{l|c}
\toprule
Model      & Space Complexity \\
\midrule
TransE & $\mathcal{O}(n_e d + n_r d)$     \\ 
TTransE & $\mathcal{O}(n_e d + n_r d + n_{\tau} d)$     \\ 
HyTE & $\mathcal{O}(n_e d + n_r d + n_{\tau} d)$     \\ 
TA-DistMult & $\mathcal{O}(n_e d + n_r d + n_{token} d)$     \\ 
ATiSE & $\mathcal{O}(n_e d + n_r d)$     \\ 
DE-SimplE & $\mathcal{O}(n_e d + n_r d)$     \\ 
TeRo & $\mathcal{O}(n_e d + n_r d + n_{\tau} d)$     \\ 
\midrule
RotateQVS & $\mathcal{O}(n_e d + n_r d + n_{\tau} d)$     \\ 
\bottomrule
\end{tabular}
}
\caption{Space complexity comparison of our RotateQVS with several baselines.}
\label{table:complexity}
\end{table}

\subsection{Theoretical Comparison Against TeRo}
\label{app:defect_TeRo}
TeRo \cite{xu2020tero} is the main baseline for our model.
The rotated head entity embedding and tail entity embedding of TeRo in complex number system are
$\vects \circ \pmb{\tau}$, and
$ \vecto \circ \pmb{\tau}$
respectively, where $\circ$ denotes Hermitian dot product. The translational score function of TeRo $
    f(s, r, o, t) =  \vert \vert\vects_{t} + \vectr - \vect{\overline{o_{t}}} \vert \vert
$
is to make
\begin{equation} \label{tero_evolution}
\vects \circ \pmb{\tau} + \vectr = \overline{ \vecto \circ \pmb{\tau}} = \overline{\pmb{\tau}} \circ \overline{\vecto} = \overline{\vecto} \circ \overline{\pmb{\tau}}\, .
\end{equation}
And we further prove that TeRo can not model relations with temporal evolution by means of reduction to absurdity.\footnote{See proof in Appendix \ref{app:proof_tero}.}

To this end, taking advantages of quaternion representation, our RotateQVS can deduce further derivation:
\begin{equation}
\begin{aligned}
& \pmb{\tau} \vects \pmb{\tau}^{-1} + \vectr = \overline{ \pmb{\tau} \vecto \pmb{\tau}^{-1}} = \pmb{\tau} \overline{\vecto} \pmb{\tau}^{-1}\\
 & \Leftrightarrow 
  \overline{\vecto} - \vects  = \pmb{\tau}^{-1} \vectr \pmb{\tau} \, , 
\end{aligned}
\end{equation}
where time stamp embeddings and relation embeddings can be particularly extracted to analyse the influence of temporal evolution on relations, since our derivation result is independent with entity embeddings.
Above all, we demonstrate that our RotateQVS can model relations with temporal evolution while TeRo cannot.\footnote{Proof process is shown in Lemma \ref{lemma:temporal-evolution}, and case based analysis is shown in Section~\ref{tempral_case}.}

\subsection{Complexity Comparison}
\label{app:complexity_comparison}

Table~\ref{table:complexity} summarizes the space complexities of several baselines and our model.
$n_e$, $n_r$, $n_{\tau}$ and $n_{token}$ denote numbers of entities, relations, time stamps, and temporal tokens used in \cite{garcia2018learning}; and $d$ is the dimension of embeddings.
The space complexity of our RotateQVS is $\mathcal{O}(n_e d + n_r d + n_{\tau} d)$, the same as TTransE \cite{leblay2018deriving}, HyTE \cite{dasgupta2018hyte} and TeRo \cite{xu2020tero}.

\begin{table}[!t]
\centering
\resizebox{0.98\columnwidth}{!}{
\begin{tabular}{l|rrrr}
\toprule
Dataset & \multicolumn{1}{c}{ICEWS14} & ICEWS05-15 & YAGO11k & GDELT \\ \midrule
{Entities} & 7,128    & 10,488   & 10,623  & 500\\
{Relations}  & 230 & 251   & 10  & 20 \\
{Time Stamps}  & 365   & 4,017   & 70  & 366\\
{Train} & 72,826    & 386,962 & 16,408  & 2,735,685 \\
{Validation}  & 8,941 & 46,275  & 2,050  & 341,961\\
{Test}  & 8,963 & 46,092 & 2,051 & 341,961
\\
\bottomrule
\end{tabular}
}
\caption{Statistics of four experimented datasets.}
\label{table:datasets}
\end{table}

\section{Experiments}

\subsection{Benchmark Datasets}\label{section_datasets}
To evaluate our proposed Quaternion embeddings, we perform link prediction task on four commonly used TKG benchmark datasets, namely {ICEWS14}, {ICEWS05-15} \cite{garcia2018learning}, {YAGO11k} \cite{dasgupta2018hyte} and {GDELT} \cite{trivedi2017know}.\footnote{GDELT is derived from \url{https://github.com/BorealisAI/de-simple/tree/master/datasets/gdelt}, and other datasets can be downloaded from \url{https://github.com/soledad921/ATISE}.}
Table~\ref{table:datasets} summarises the details of the four datasets, where it is easy to find ICEWS14 and ICEWS05-15 have more quantitative relations than the other two datasets.

ICEWS \cite{lautenschlager2015icews} is a repository containing political events with a specific timestamp.
ICEWS14 and ICEWS05-15 \cite{garcia2018learning} are two subsets of ICWES corresponding to facts in 2014 and facts between 2005 and 2015.

YAGO11k \cite{dasgupta2018hyte} is a subset of YAGO3 \cite{Mahdisoltani2015YAGO3AK}, where time annotations are represented as time intervals.
We derive the dataset from HyTE \cite{dasgupta2018hyte} to obtain the same year-level granularity by dropping the month and date information, which results in 70 different time stamps.

For GDELT, we use the subset extracted by \citeauthor{trivedi2017know}, consisting of the facts from April 1, 2015 to March 31, 2016.
We take the same pretreatment of the train, validation and test sets as  \cite{goel2020diachronic}, to make the
problem into a TKGC rather than an extrapolation problem.

\begin{table*}[!t]
\centering
\small
\begin{tabular}{@{}lcccc|cccc@{}}
\toprule
\multirow{2}{*}{Dataset}   & \multicolumn{4}{c|}{ICEWS14}   & \multicolumn{4}{c}{ICEWS05-15} \\
\cmidrule{2-9}
   & Hits@1        & Hits@3  & Hits@10  & MRR  & Hits@1           & Hits@3  & Hits@10 & MRR \\
\midrule
TransE \cite{bordes2013translating}      & 0.094        & -  &  0.637 & 0.280  & 0.090           & - & 0.663 & 0.294   \\
DistMult \cite{Yang2015EmbeddingEA}   & 0.323        & - & 0.672 & 0.439  & 0.337           & - & 0.691  & 0.456  \\
RotatE \cite{SunDNT19}    & 0.291        & 0.478 & 0.690 & 0.418 &  0.164          & 0.355 & 0.595 & 0.304 \\
QuatE \cite{zhang2019quaternion}      &  0.353       & 0.530 & 0.712 & 0.471 & 0.370           & 0.529 & 0.727 & 0.482  \\
\midrule
TTransE  \cite{leblay2018deriving} & 0.074        & - & 0.601 & 0.255 & 0.084           & - & 0.616 & 0.271  \\
HyTE \cite{dasgupta2018hyte}       & 0.108        & 0.416 & 0.655 & 0.297 & 0.116           & 0.445 & 0.681 & 0.316 \\
TA-DistMult \cite{garcia2018learning} & 0.363        & - & 0.686 & 0.477 &  0.346          & - & 0.728 & 0.474  \\
ATiSE \cite{xu2019temporal}      & 0.436        & \underline{0.629} & \underline{0.750} & 0.550 &  0.378          & 0.606 & 0.794 & 0.519 \\
DE-SimplE \cite{goel2020diachronic}  & 0.418        & 0.592 & 0.725 & 0.526 &  0.392          & 0.578 & 0.748 & 0.513  \\
\midrule
TeRo \cite{xu2020tero}       &  0.468       & 0.621 & 0.732 & 0.562 &  0.469          & 0.668 & 0.795 & 0.586   \\
TeRo-Large       &  0.432     & 0.596 & 0.722 & 0.534 &  0.395          & 0.627 & 0.800 & 0.534 \\
\midrule
RotateQVS-Small (ours) & \underline{0.489}       &0.625 & 0.737 & \underline{0.575} & \underline{0.473}         & \underline{0.685} & \underline{0.802} & \underline{0.591} \\
RotateQVS (ours) & \bfseries 0.507        &\bfseries 0.642 &\bfseries 0.754 &\bfseries 0.591 &\bfseries  0.529          &\bfseries 0.709 &\bfseries 0.813 &\bfseries 0.633  \\
\midrule
   & \multicolumn{4}{c|}{YAGO11k}   & \multicolumn{4}{c}{GDELT} \\
\cmidrule{2-9}
   &  Hits@1        & Hits@3  & Hits@10  & MRR & Hits@1        & Hits@3  & Hits@10  & MRR \\
\midrule
TransE      & 0.015 & 0.138 & 0.244 &0.100  & 0.0   & 0.158 & 0.312 & 0.113  \\
DistMult   & 0.107 & 0.161 & 0.268 & 0.158  & 0.117 & 0.208 & 0.348  & 0.196 \\
RotatE       &  0.103  & 0.167 & 0.305 & 0.167         &-    & - & - & -\\
QuatE      &  0.107  & 0.148 & 0.270 & 0.164 &-    & - & - & -\\
\midrule
TTransE   & 0.020 & 0.150 & 0.251 & 0.108 & 0.0           & 0.160 & 0.318 & 0.115\\
HyTE       &  0.015  & 0.143 & 0.272 & 0.105 
& 0.0           & 0.165 & 0.326 & 0.118 \\
TA-DistMult  & 0.103 &0.171 &0.292 & 0.161 & 0.124          & 0.219 & 0.365 & 0.206 \\
ATiSE      & 0.110  & 0.171 & 0.288 & 0.170 &-    & - & - & -\\
DE-SimplE  & -  & -& - & - &  0.141 &0.248 & 0.403 & 0.230  \\
\midrule
TeRo       & \underline{0.121} &\underline{0.197} &0.319 &\underline{0.187} & 0.154   & 0.264  & 0.420  & 0.245  \\
TeRo-Large       & 0.094 & \bfseries 0.199 & \bfseries 0.323 & 0.181 &  0.163          & \underline{0.278} & \underline{0.437} & 0.256 \\
\midrule
RotateQVS-Small & \bfseries 0.124       & 0.193 & 0.320 & \underline{0.187} & \underline{0.165}        & 0.270 & 0.428 & \underline{0.259}\\
RotateQVS & \bfseries 0.124 & \bfseries 0.199 & \bfseries 0.323 & \bfseries0.189 &\bfseries  0.175    &\bfseries 0.293 &\bfseries 0.458 &\bfseries 0.270\\
\bottomrule
\end{tabular}

\caption{Results on link prediction task over four experimented datasets. The best score is in \textbf{bold} and second best score is \underline{underlined}. 
}
\label{table:results}
\end{table*}

\subsection{Evaluation Protocol}
Link prediction task that aims to infer incomplete time-wise fact with a missing entity ($(s, r, ? , t)$ or $(?, r, o , t)$) is adopted to evaluate the proposed model.
During the inference, we follow the same procedure of \citeauthor{xu2020tero} to generate candidates. For a test sample $(s, r, o , t)$, we first generate candidate quadruples set $C = \{(s, r, \overline{o}, t) : \overline{o} \in \mathcal{E}\} \cup \{(\overline{s}, r, o, t) : \overline{s} \in \mathcal{E}\}$ by replacing $s$ or $o$ with all possible entities, and then rank all the quadruples by their scores (Equation~\ref{equation:score}) under the time-wise filtered settings \cite{xu2019temporal,goel2020diachronic}.

The performance is reported on the standard evaluation metrics: the proportion of correct triples ranked in top 1, 3 and 10 (Hits@1, Hits@3, and Hits@10), and Mean Reciprocal Rank (MRR).
All the metrics (Hits@1, Hits@3, Hits@10 and MRR) are the higher the better.
For all experiments, we report averaged results across 5 runs, and we omit the variance as it is generally low.

\subsection{Baselines}
\label{sec:Baselines}
We compare with both sota static and temporal KGE baselines.
For static baselines, we use TransE \cite{bordes2013translating}, DistMult \cite{Yang2015EmbeddingEA}, RotatE \cite{SunDNT19}, and QuatE \cite{zhang2019quaternion}.
For TKGE methods, we consider TTransE \cite{leblay2018deriving}, HyTE \cite{dasgupta2018hyte}, TA-DistMult \cite{garcia2018learning}, DE-SimplE \cite{goel2020diachronic}, ATiSE \cite{xu2019temporal}, and TeRo \cite{xu2020tero}.\footnote{See complexity comparison in Appendix \ref{app:complexity_comparison}.}

Note that TeRo \cite{xu2020tero} is also based on the idea of rotations, and thus we consider TeRo as a directly baseline.
Because our quaternion representation ($\vect{a} + \vect{b} \vecti +\vect{c} \vectj +\vect{d} \vectk$) doubles the embedding parameters of TeRo which uses complex representation ($\vect{a} + \vect{b} \vecti$),  we further adopt two models for fair comparisons:
(i) TeRo-Large: TeRo using double embedding dimension;\footnote{We reuse the original implementation of \cite{xu2020tero} from \url{https://github.com/soledad921/ATISE} and follow the their best setups.}
(ii) RotateQVS-Small: The proposed RotateQVS with half embedding dimension.
By doing so, their parameter complexities can be comparable with TeRo's.

\subsection{Results}
The experimental results over four TKG datasets are shown in Table~\ref{table:results}.\footnote{See hyperparameter setup in Appendix \ref{app:hyperparameter}.}
Overall, TKGE methods are better than static KGE methods, which shows the effectiveness of modeling temporal information.
For the proposed RotateQVS, we observe that our model outperforms all the baseline models over the four datasets across all metrics consistently.\footnote{We also take time granularity analysis and embedding dimension analysis in Appendix \ref{app:time_granularity} and \ref{app:dimension}.}
To demonstrate the superiority of the proposed quaternion method, we compare our RotateQVS with the direct baseline TeRo \cite{xu2020tero}.
For fair comparisons of model sizes, we observe that our RotateQVS outperforms TeRo-Large and RotateQVS-Small outperforms TeRo.
This shows our methods with quaternion embeddings makes great improvements, demonstrating our advantages.
Specially, we see that our RotateQVS achieves more improvements on ICEWS14 and ICEWS05-15 datasets.
We believe this is because these two datasets have much more quantitative relations (see Table~\ref{table:datasets}) and it is also evident our method behaves better on datasets with complex relation patterns.

\subsection{Analysis and Case Study}
To further demonstrate the learned quaternion embeddings and the ability of our model, we perform case studies on multiple relation patterns, through visualization and quantitative analysis on intuitive examples from ICEWS14.

\subsubsection{Symmetric/Asymmetric/Inversion Patterns}
Since symmetric, asymmetric and inversion patterns have been discussed in previous work \cite{SunDNT19,xu2020tero}, we present the case studies of them to Appendix~\ref{app:case_study_3}.

\subsubsection{Temporal-evolution Pattern} \label{tempral_case}

\begin{table*}[t]
\centering
\resizebox{0.95\textwidth}{!}{
\begin{tabular}{lc|c|c|c|c|c}
\toprule
 & & \textbf{Head entity} & \textbf{Relation}  & \textbf{Tail entity} & \textbf{Time} & \textbf{Similarity}
\\ \midrule
\multirow{3}{*}{\rotatebox{90}{\footnotesize{\textbf{Example 1}}}} & Base fact & \multirow{3}{*}{\Large{John Kerry}}
&  Express intent to meet or negotiate  &  \multirow{3}{*}{\Large{Pietro Parolin}}  &  2014-01-13 & \multirow{2}{*}{0.810} \\
\cline{2-2} \cline{4-4} \cline{6-6}
& True fact &  &  Consult  &   &  \multirow{2}{*}{2014-01-16} \\ \cline{2-2} \cline{4-4} \cline{7-7}
& Negative &  &  Detonate nuclear weapons  &    &   &0.508 \\
\midrule
\multirow{3}{*}{\rotatebox{90}{\footnotesize{\textbf{Example 2}}}} & Base fact & \multirow{3}{*}{\Large{Member of Legislative (Govt) (Iran)}}   & Make statement &  \multirow{3}{*}{\Large{Iran}}    & 2014-03-16 & \multirow{2}{*}{0.819} \\ \cline{2-2} \cline{4-4} \cline{6-6}
& True Fact &  &  Make statement   &   &  \multirow{2}{*}{2014-05-04} \\ \cline{2-2} \cline{4-4} \cline{7-7}
& Negative &  &  Detonate nuclear weapons  &    &    &0.492 \\
\midrule 
\multirow{3}{*}{\rotatebox{90}{\footnotesize{\textbf{Example 3}}}} & Base fact & \multirow{3}{*}{\Large{Federal Bank}}  &  Make a visit &  \multirow{3}{*}{\Large{European Central Bank}}     &2014-02-04 & \multirow{2}{*}{0.815} \\ \cline{2-2} \cline{4-4} \cline{6-6}
& True fact &  &  Make statement  &   &  \multirow{2}{*}{2014-02-25}    \\ \cline{2-2} \cline{4-4} \cline{7-7}
& Negative &  &  Receive inspectors  &    &   &0.510 \\
\bottomrule
\end{tabular}
}
\caption{Examples of temporal-evolution patterns in ICEWS14 dataset. The similarity score is based on base fact.}
\label{table:temporal-evolution}
\vspace{-0.1in}
\end{table*}

As shown in Lemma \ref{lemma:temporal-evolution}, if a relation $r_1$ and a relation $r_2$ are evolving over time from $t_1$ (time stamp $\tau_1$) to $t_2$ (time stamp $\tau_2$), we have
\begin{equation} \label{equation:temporal-evolution-pattern}
    \pmb{\tau}_2 \pmb{\tau}_1^{-1} \vect{r}_1 (\pmb{\tau}_2 \pmb{\tau}_1^{-1})^{-1} = \vect{r}_2.
\end{equation}

To analyse the temporal-evolution pattern, we focus on the relations between the same head and tail entities with different time stamps.
For example, from ICEWS14, we observe a base fact \textit{(South Korea, Engage in negotiation, North Korea, 2014-02-12)} and a temporal-evolution fact \textit{(South Korea, Sign formal agreement, North Korea, 2014-02-15)}, where \exam{Sign formal agreement} is considered as the consequence of \exam{Engage in negotiation}.
Thus, in our model, the embeddings of \textit{Sign formal agreement} at time stamp \textit{2014-02-15} and of \textit{Engage in negotiation} at \textit{2014-02-12} should satisfy Equation~ \ref{equation:temporal-evolution-pattern}.

\begin{figure}[!t]
\centering
\includegraphics[width=0.9\columnwidth]{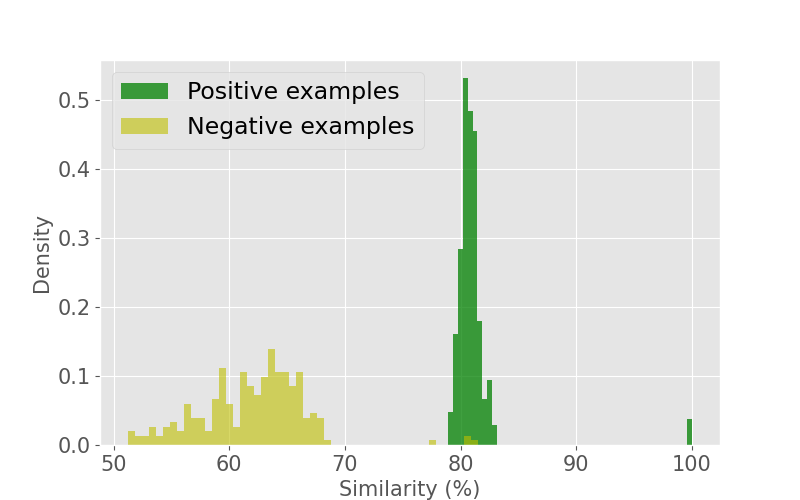} 
\caption{Density histogram with bin size 1\% of similarity scores for temporal-evolution relations. All positive and negative examples are randomly sampled and compared with the base relation \exam{Engage in negotiation}.}
\label{figure:temporal_examples}
\end{figure}

To illustrate this pattern, we measure the matrix cosine similarity between $\vect{r}_2$ (base) and $\pmb{\tau}_2 \pmb{\tau}_1^{-1} \vect{r}_1 (\pmb{\tau}_2 \pmb{\tau}_1^{-1})^{-1}$ (temporal-evolved).
For each true fact, we sample a random negative relation and show their similarity difference.
Figure~\ref{figure:temporal_examples} illustrates the density histogram of similarities with random 250 fact quadruples at different time stamps between \textit{South Korea} and \textit{North Korea}.
We observe that the distributions of positive examples and negative examples are distinct, which explains our RotateQVS can model temporal-evolution patterns more effectively.
Comparing with TeRo \cite{xu2020tero}, which is the main baseline for our model, we show TeRo cannot model this pattern theoretically (see Section \ref{app:defect_TeRo}).

\begin{figure}[!t]
  \centering
    \subfigure[$Re(\vect{r_1}) - Re(\vect{r_2}) = \vect{0}$]{\includegraphics[width=0.23\textwidth]{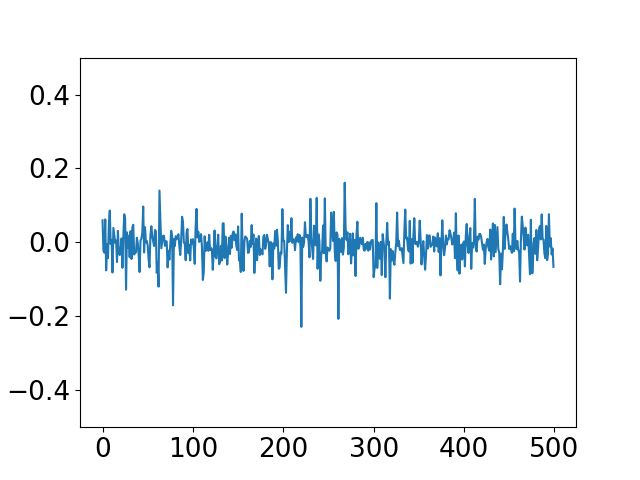}} 
	\subfigure[$\vert \vert\vect{r_1}\vert \vert - \vert \vert \vect{r_2}\vert \vert = \vect{0}$]{\includegraphics[width=0.23\textwidth]{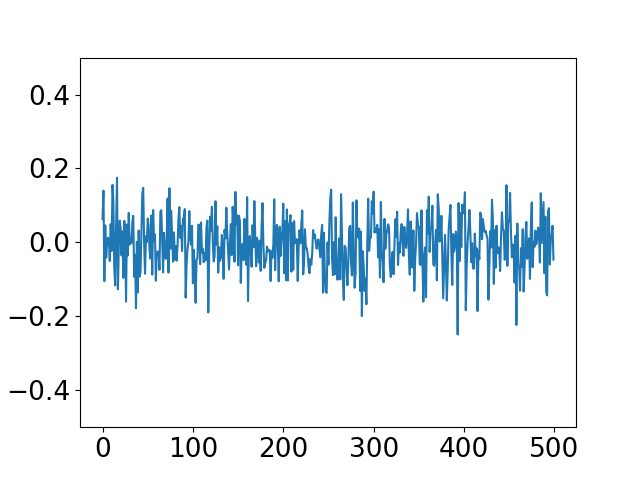}} \\
  \caption{Quaternion representations of Equation \ref{equation:further_deductions} for temporal-evolution pattern.}
  \label{figure:tp_deduction}
\end{figure}

In addition, Figure~\ref{figure:tp_deduction} shows our quaternion representation do well in reflecting Equation \ref{equation:further_deductions}, the sufficient and unnecessary deductions of theoretical analysis for temporal-evolution pattern.

More examples of temporal-evolution pattern are shown in Table~ \ref{table:temporal-evolution}, where we use the relation in base fact and time information to get a generated embedding $\pmb{\tau_2} \pmb{\tau_1}^{-1} \vect{r_1} (\pmb{\tau_2} \pmb{\tau_1}^{-1})^{-1}$, and also sample a random negative relation for each example.
We compute the matrix cosine similarity between $\pmb{\tau_2} \pmb{\tau_1}^{-1} \vect{r_1} (\pmb{\tau_2} \pmb{\tau_1}^{-1})^{-1}$ and $\vect{r_2}$, and also compute the similarity between $\pmb{\tau_2} \pmb{\tau_1}^{-1} \vect{r_1} (\pmb{\tau_2} \pmb{\tau_1}^{-1})^{-1}$ and the embedding of another relation in the negative sample.
Time stamps in negatives are taken as same as the true facts.
The comparison between the two sets of results can once again prove the ability of our model in modeling this pattern.

\subsection{Convergence Analysis}
\begin{figure}[t!]
\centering
\includegraphics[width=0.9\columnwidth]{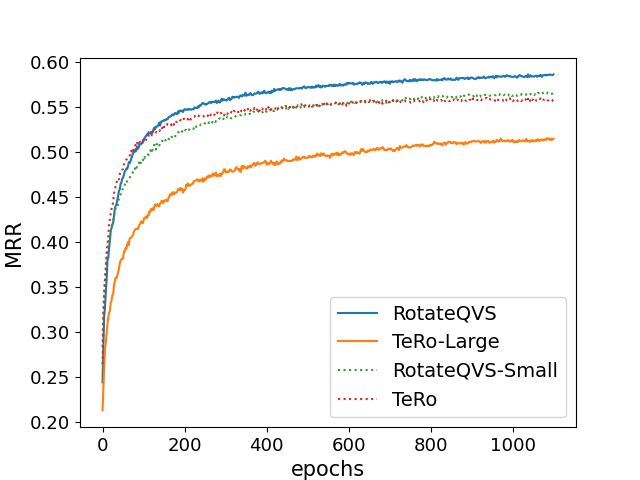} 
\caption{The convergence study of RotateQVS, TeRo-Large, RotateQVS-Small and TeRo by epochs on ICEWS14 test set, and we use the metric MRR here.}
\label{figure:convergence}
\vspace{-0.1in}
\end{figure}

For convergence analysis, we consider two fair comparisons, where the compared two methods have the same number of parameters:\footnote{Refer to Section \ref{sec:Baselines} for more details} RotateQVS (\textcolor{myblue}{blue solid} line) vs. TeRo-Large (\textcolor{myyellow}{yellow solid} line) and RotateQVS-Small (\textcolor{mygreen}{green dotted} line) vs. TeRo (\textcolor{myred}{red dotted} line) in Figure~\ref{figure:convergence}.
We observe that RotateQVS and TeRo-Large converge at approximately the same rate, and so do RotateQVS-Small and TeRo.
We can conclude that our proposed RotateQVS can achieve better results in comparisons of both large and small levels without sacrificing additional training efforts.

\section{Related work}

Models working on Static Knowledge graph have been well studied \cite{crosse,XuRKKA20,MaoWXLW20,BISC} with semantic and structure information. Translation based methods, e.g. TransE \cite{bordes2013translating} and TransR \cite{linlearning}, formalise the factual distance between a head entity $s$ and a tail entity $o$ with the translation carried out by the relation. Adopting tensor factorization with a bi-linear transformation, semantic matching models, e.g. RESCAL \cite{nickel2013tensor} and DistMult \cite{Yang2015EmbeddingEA}, capture the semantic relevance of entities. Recently, more attention were paid to study various relation patterns. RotatE \cite{SunDNT19} treat each relation as a rotation so that symmetric/asymmetric, inversion and composition patterns can be inferred to predict missing links. Further, quaternion number system \cite{Hamilton1844} is applied to model more complex composition patterns in 3D space, such as Rotate3D \cite{gao2020rotate3d} and QuatE \cite{zhang2019quaternion}.

Many aforementioned methods \cite{dasgupta2018hyte,leblay2018deriving,trivedi2017know,garcia2018learning,goel2020diachronic,ChronoR} are extended from static Static KGs to TKGs. They integrate time information into previous static methods as independent features. Others study the dynamic evolution of TKG. ATiSE \cite{xu2019temporal} regards the temporal evolution of entity and relation embeddings as combinations of trend component, seasonal component and random component.
CyGNet \cite{ZhuCFCZ21} proposes a time-aware copy-generation mechanism leveraging known facts in the past to predict unknown facts in the future.
TeRo \cite{xu2020tero} defines the temporal evolution of entity embedding as a rotation in the complex vector space.
Inspired by TeRo, our RotateQVS further represents temporal entities as rotations in quaternion vector space and obtains more advantages.\footnote{Refer to Section \ref{app:defect_TeRo} for more details.}

Modeling various temporal relation patterns \cite{goel2020diachronic,xu2020tero}, especially the temporal-evolution patterns, is crucial for TKGE and the following TKGC.
\citeauthor{zhang2021spatial} mentions the time-evolution property, but does not make a systematic research on it.
It remains an open research question with few researches.
Our work (RotateQVS) takes inspirations from the idea of rotation and generalises it into the quaternion number system to model the complex temporal-evolution pattern that TeRo can hardly do.

\section{Conclusion}
In this paper, we introduce a novel TKGC method RotateQVS which represents temporal information of TKGs as rotations in quaternion vector space. Targeting temporal interpretability, we theoretically analyse that RotateQVS can model various relation patterns and demonstrate it with extensive experiments.
Compared to previous methods, RotateQVS has made significant improvements on link prediction tasks over four benchmark datasets.
Furthermore, we show our RotateQVS has great advantages in modeling various relation patterns with temporal evolution.

\section*{Acknowledgements}
This work is supported by the Key R\&D Program of Guangdong  Province (No.2019B010136003), the National Natural Science Foundation of China (No. 61732004, 61732022).

\bibliography{reference}
\bibliographystyle{acl_natbib}

\clearpage
\appendix

\section{Proof of Theorem \ref{Theorem_rotation}}
\label{app:proof_theorem1}
\begin{figure}[!t]
\centering
\includegraphics[width=0.85\columnwidth]{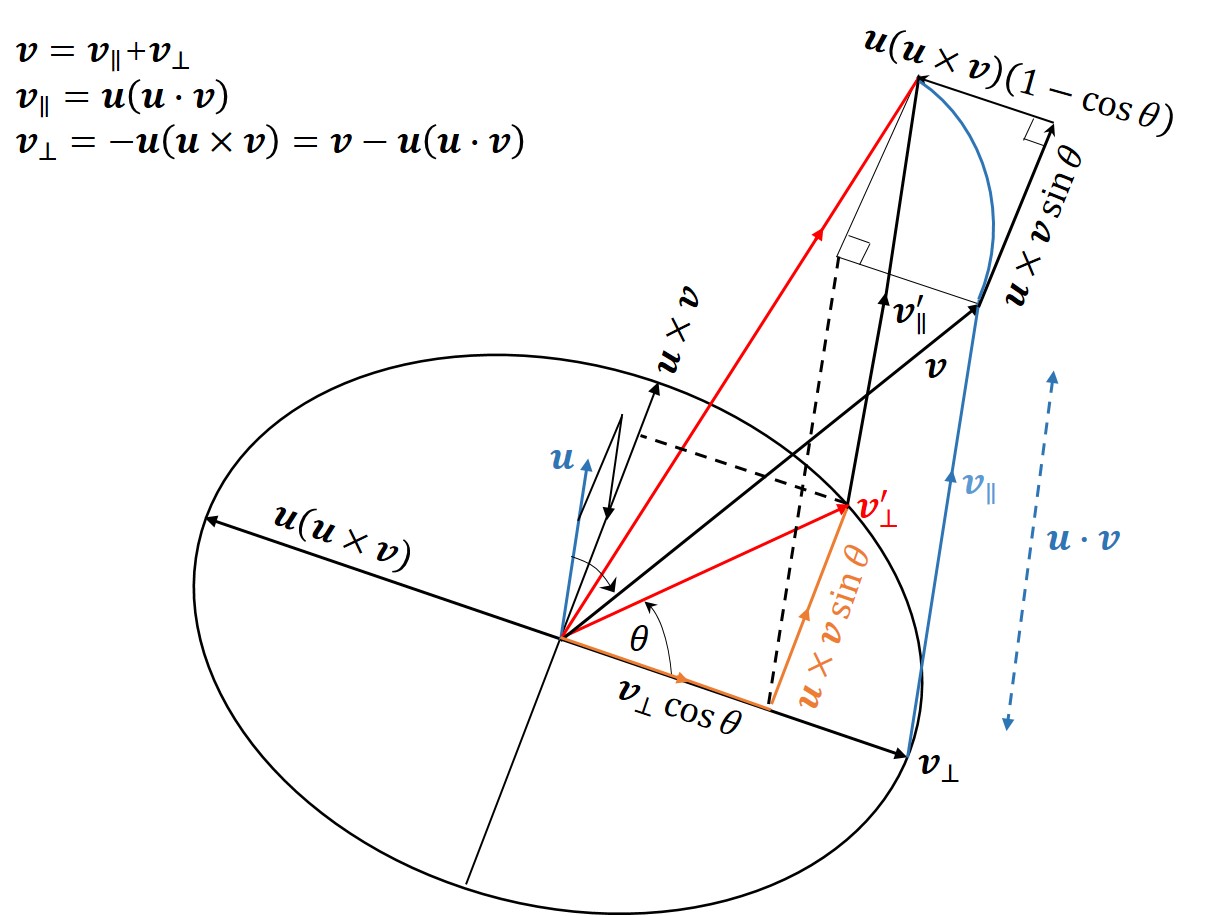} 
\caption{An illustration of our rotation equation, which satisfies the Rodrigues' rotation formula \cite{rodrigues1840lois}, where $\vect{v'}$ is the result of vector $\vect{v}$ rotating $\theta$ around rotation axis $\vect{u}$.}
\label{figure:Fodrigues}
\end{figure}

\begin{proof}
Based on Equation \ref{equation:inverse_q}, for a unit quaternion $q$, it follows that $q^{-1} = \overline{q}$.
Unfolding the Equation \ref{equation:rotation}, we can get
\begin{align}
    \vect{v'} 
    &= (\cos \frac{\theta}{2} + \vect{u} \sin \frac{\theta}{2}) \vect{v} (\cos \frac{\theta}{2} - \vect{u} \sin \frac{\theta}{2}) \nonumber\\
    &=\vect{v} {\cos}^2 \frac{\theta}{2} + (\vect{u} \vect{v} - \vect{v} \vect{u}) \sin \frac{\theta}{2} \cos \frac{\theta}{2} \nonumber\\
    &\quad - \vect{u} \vect{v} \vect{u} {\sin}^2 \frac{\theta}{2} \, .
\end{align}

Bearing in mind that $\vect{u} \vect{v} = \vect{u} \times \vect{v} - \vect{u} \cdot \vect{v}$ (based on the Equation \ref{equation:vector_multiplication}), further we can get
\begin{align}
    \vect{v'} 
    &=\vect{v} {\cos}^2 \frac{\theta}{2} + 2(\vect{u} \times \vect{v}) \sin \frac{\theta}{2} \cos \frac{\theta}{2}  
    \nonumber \\
    &\quad  - ((\vect{u} \times \vect{v}) - (\vect{u} \cdot \vect{v}))\vect{u} {\sin}^2 \frac{\theta}{2} 
    \nonumber\\
    &=\vect{v} ({\cos}^2 \frac{\theta}{2} - {\sin}^2 \frac{\theta}{2}) + (\vect{u} \times \vect{v}) (2\sin \frac{\theta}{2} \cos \frac{\theta}{2}) \nonumber\\
    & \quad +
    \vect{u} (\vect{u} \cdot \vect{v}) ({2\sin}^2 \frac{\theta}{2}) \, .
\end{align}

Using trigonometric identities, we can get
\begin{align}\label{equation:our_formula}
    \vect{v'}
    &=\vect{v} \cos \theta  + (\vect{u} \times \vect{v}) \sin \theta \nonumber\\
    & \quad + \vect{u} (\vect{u} \cdot \vect{v}) (1 - \cos \theta) \nonumber\\
    &=(\vect{v} - \vect{u} (\vect{u} \cdot \vect{v}))\cos \theta  + (\vect{u} \times \vect{v}) \sin \theta \nonumber\\
    & \quad + \vect{u} (\vect{u} \cdot \vect{v}) \nonumber\\
    &=\vect{{v}_{\perp}} \cos \theta  + (\vect{u} \times \vect{v}) \sin \theta +
    \vect{{v}_{\parallel}}
\end{align}
where $\vect{{v}_{\perp}} = \vect{v} - \vect{u} (\vect{u} \cdot \vect{v})$ and $\vect{{v}_{\parallel}} = \vect{u} (\vect{u} \cdot \vect{v})$ are the components of $\vect{v}$ (perpendicular and parallel to the axis $\vect{u}$ respectively).
Our Equation~\ref{equation:our_formula} satisfies the Rodrigues' rotation formula \cite{rodrigues1840lois} in 3D vector space (illustrated in Figure~\ref{figure:Fodrigues}).
Therefore, the Equation~\ref{equation:rotation} is proven to be a rotation in 3D vector space.
\end{proof}

\section{Proof of Lemma \ref{lemma:symmetric}}
\label{app:proof_symmetric}
\begin{proof}
For symmetric pattern, considering our rotation based temporal representation, we express the $ r(s,o,t) \land r(o,s,t)$ in Definition \ref{definition:r1} as:
\begin{equation}
\begin{aligned}
\left\{  
     \begin{aligned}
     \overline{\vecto} - \vects  = \pmb{\tau}^{-1} \vectr \pmb{\tau} \\
    \overline{\vects} - \vecto  = \pmb{\tau}^{-1} \vectr \pmb{\tau}
    \end{aligned}
\right.
\Leftrightarrow 
\vectr + \overline{\vectr}=0
\Leftrightarrow 
Re(\vectr)=0\,,
\end{aligned}
\end{equation}
where $Re$ denotes the real part of a quaternion representation.
\end{proof}

\section{Proof of Lemma \ref{lemma:asymmetric}}
\label{app:proof_asymmetric}
\begin{proof}
For asymmetric pattern, $ r(s,o,t) \land \lnot r(o,s,t)$ in Definition \ref{definition:r2} can be expressed as:
\begin{equation}
\begin{aligned}
&\left\{  
     \begin{aligned}
     \overline{\vecto} - \vects  = \pmb{\tau}^{-1} \vectr \pmb{\tau} \\
    \overline{\vects} - \vecto  \neq \pmb{\tau}^{-1} \vectr \pmb{\tau}
    \end{aligned}
\right.
\Leftrightarrow 
\vectr + \overline{\vectr}\neq0
\Leftrightarrow 
Re(\vectr)\neq0 \, .
\end{aligned}
\end{equation}
\end{proof}

\section{Proof of Lemma \ref{lemma:inversion}}
\label{app:proof_inversion}
\begin{proof}
For inversion pattern, $ r_1(s,o,t) \land r_2(o,s,t)$ in Definition \ref{definition:r3} can be expressed as:
\begin{equation}\label{equation:inverse}
\begin{aligned}
&\left\{  
     \begin{aligned}
     \overline{\vecto} - \vects  = \pmb{\tau}^{-1} \vect{r}_1 \pmb{\tau} \\
    \overline{\vects} - \vecto  = \pmb{\tau}^{-1} \vect{r}_2 \pmb{\tau}
    \end{aligned}
\right.
\\
&\Leftrightarrow
\vect{r}_1 + \overline{\vect{r}_2}=0 \quad
\Leftrightarrow 
\left\{  
     \begin{aligned}
     &Re(\vect{r}_1) + Re(\vect{r}_2) = \vect{0} \\
    &Im(\vect{r}_1) - Im(\vect{r}_2) = \vect{0} \,,
    \end{aligned}
\right.
\end{aligned}
\end{equation}
where $Im$ denotes the imaginary part of a quaternion representation.
\end{proof}

\section{Statistics of several previous TKGE methods modeling on various relation patterns}
\label{app:Statistics}

Table \ref{table:relations} shows the statistics of several previous TKGE methods modeling on various relation patterns, containing symmetry, asymmetry, inversion and temporal-evolution. 

\begin{table}[t!]
\centering
\resizebox{0.98\columnwidth}{!}{
\begin{tabular}{l|c|c|c|c}
\toprule
Methods     & Symmetry & Asymmetry & Inversion & Temporal-evolution \\
\midrule
TTransE   & \XSolid   & \Checkmark       & \Checkmark & \XSolid \\
TA-DistMult & \Checkmark  & \XSolid       & \XSolid & \XSolid \\
DE-SimplE & \Checkmark    & \Checkmark  & \Checkmark     & \XSolid \\ 
TeRo & \Checkmark    & \Checkmark  & \Checkmark     & \XSolid \\ 
RotateQVS (ours) & \Checkmark    & \Checkmark  & \Checkmark     & \Checkmark \\
\bottomrule
\end{tabular}
}
\caption{Statistics of several previous TKGE methods modeling on various relation patterns.}
\label{table:relations}
\end{table}

\section{Proof by Contradiction for TeRo}
\label{app:proof_tero}
\begin{proof}
Supposing TeRo \cite{xu2020tero} can model the temporal-evolution relation pattern (defined in Definition \ref{definition:r4}), then relations with temporal-evolution pattern will satisfy the Equation \ref{tero_evolution}.
Notice that our relation patterns defined are unconcerned with some specific entities, but focusing on the general rules among relations inside the universal entities.

If relation $r_1$ and $r_2$ are evolving over time from $t_1$ to $t_2$, considering the same head entity $s$ and two different tail entities $o_1$ and $o_2$ which satisfy $ r_1(s,o_1,t_1) \land r_2(s,o_1,t_2)$ and $ r_1(s,o_2,t_1) \land r_2(s,o_2,t_2)$, we can get
\begin{equation}\small
\begin{aligned}
&\left\{  
     \begin{aligned}
     &\vects \circ \pmb{\tau}_1 + \vect{r}_1 = \overline{\vect{o}_1} \circ \overline{\pmb{\tau}_1}
     \\
    &\vects \circ \pmb{\tau}_2 + \vect{r}_2 = \overline{\vect{o}_1} \circ \overline{\pmb{\tau}_2}
    \end{aligned}
\right.
\hspace{0.5em}
\land
\hspace{0.5em}
\left\{  
     \begin{aligned}
     &\vects \circ \pmb{\tau}_1 + \vect{r}_1 = \overline{\vect{o}_2} \circ \overline{\pmb{\tau}_1}
     \\
    &\vects \circ \pmb{\tau}_2 + \vect{r}_2 = \overline{\vect{o}_2} \circ \overline{\pmb{\tau}_2}
    \end{aligned}
\right. \,, \\
\end{aligned}
\end{equation}
where we can find the derivations depend on entity embeddings. And we can further derive that
\begin{equation}\small
\begin{aligned}
&\left\{  
     \begin{aligned}
     &\vects \circ (\pmb{\tau}_2 - \pmb{\tau}_1) + (\vect{r}_2 - \vect{r}_1) = \overline{\vect{o}_1} \circ (\overline{\pmb{\tau}_2} - \overline{\pmb{\tau}_1})
     \\
     &\vects \circ (\pmb{\tau}_2 - \pmb{\tau}_1) + (\vect{r}_2 - \vect{r}_1) = \overline{\vect{o}_2} \circ (\overline{\pmb{\tau}_2} - \overline{\pmb{\tau}_1})
    \end{aligned}
\right. \\
& \Leftrightarrow
 \vect{o}_1 =  \vect{o}_2 \,,
\end{aligned}
\end{equation}
where two different tail entities $o_1$ and $o_2$ have the exactly same embeddings in TeRo. Obviously, it is not in line with our common sense and has a big problem in modelling the temporal-evolution relation pattern.
\end{proof}

\section{Hyperparameter}
\label{app:hyperparameter}
To seek and find proper hyperparameters, we utilize a grid search empirically over the following ranges for all three datasets: embedding dimension in $\{100, 200, 300, 400, 500\}$, learning rate in $\{1, 0.5, 0.3, 0.1, 0.05, 0.03, 0.02, 0.01\}$, the ratio of negative over positive training sample in $\{1, 3, 5, 10\}$, the margin used in loss function in $\{1, 2, 3, 5, 10, 20, \cdots, 120\}$,  the time granularity parameter in $\{1, 2\}$, and the optimizer we use is Adagrad.

And we have found out the best hyperparameters combination as follows: for ICEWS14, we set the margin as 110, the time granularity parameter as 1; for ICEWS05-15, we set the margin as 120, the time granularity parameter as 2; for YAGO11k, we set the margin as 50, the time granularity parameter as 100; for GDELT, we set the margin as 110, the time granularity parameter as 1; and for all the datasets, we choose the learning rate as 0.1, the embedding dimension as 500, the ratio of negative over positive training sample as 10.

\section{Time Granularity Analysis}
\label{app:time_granularity}
\begin{figure}[!t]
\centering
\includegraphics[width=0.95\columnwidth]{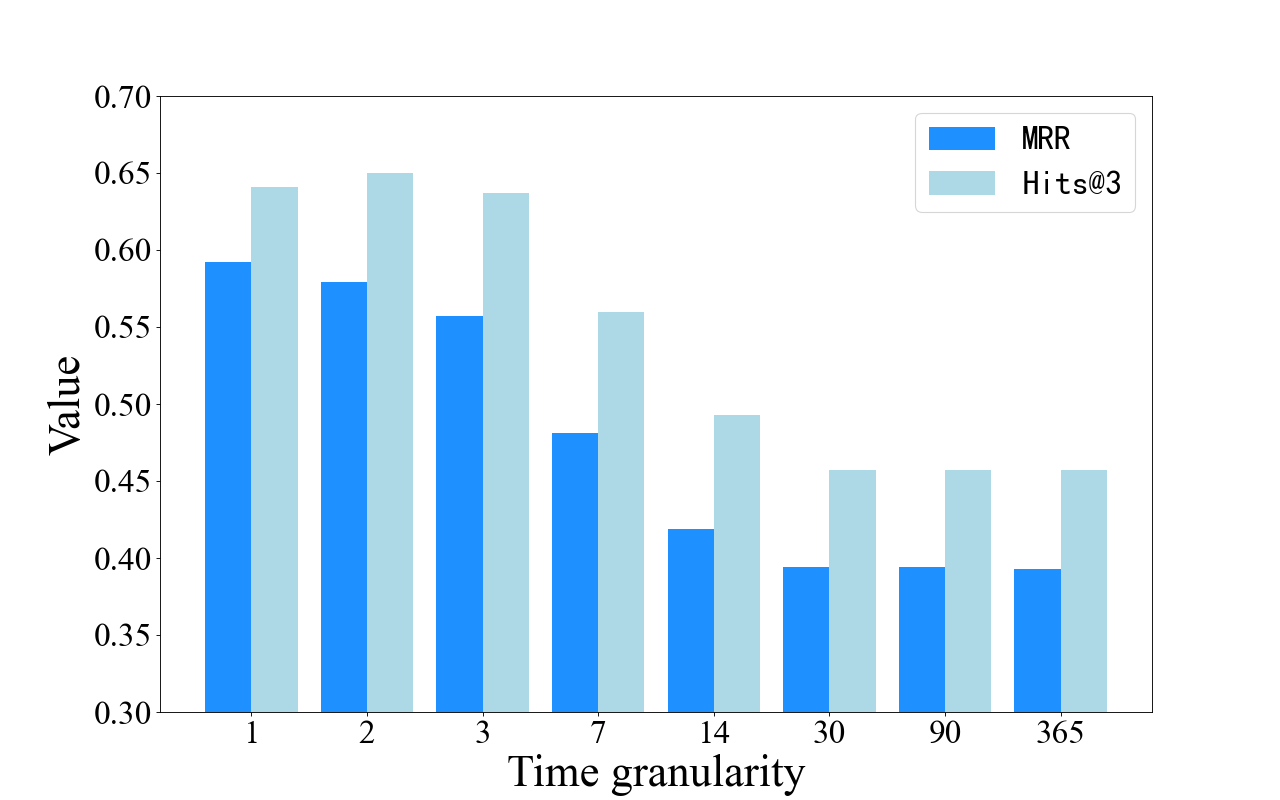} 
\caption{Results of RotateQVS with different time granularities on ICEWS14 dataset.}
\label{figure:granularity}
\end{figure}

As shown in Figure \ref{figure:granularity}, we take time granularity analysis on ICEWS14 dataset.
It find that the results of smaller granularities are better than that of larger granularities, as larger-granularity setups fuzz the time information.
Smaller granularity means more time stamps to compute, while we believe in current dataset the number of time-stamps are relatively small compared with the numbers of relations and entities, and thus we suggest small time granularity in TKG tasks.

\section{Size of Embedding Dimension}
\label{app:dimension}

\begin{figure}[!t]
\centering
\includegraphics[width=0.95\columnwidth]{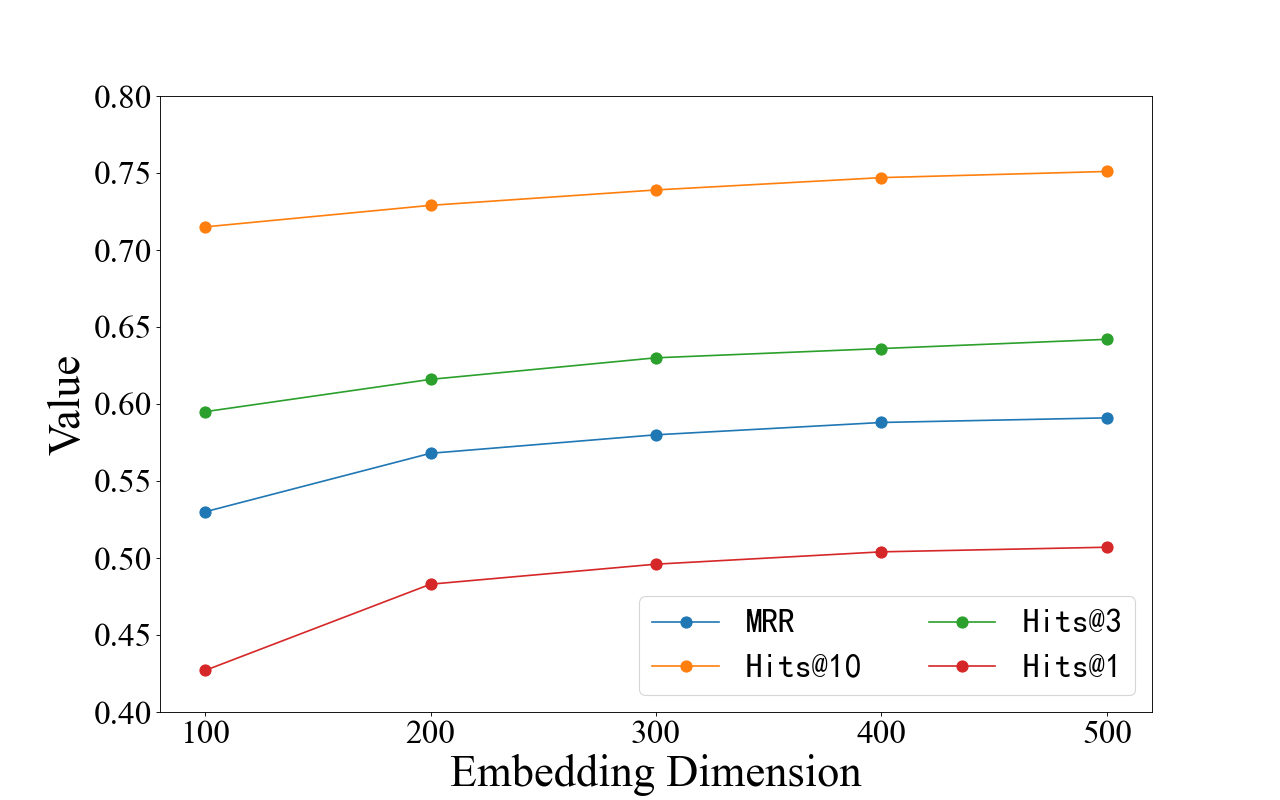} 
\caption{Results of RotateQVS with different embedding dimensions on ICEWS14 dataset.}
\label{figure:dimension}
\end{figure}

As shown in Figure \ref{figure:dimension}, we take embedding dimension analysis on ICEWS14 dataset. We can find out that the values of all the four metrics increase as the dimension increases from 100 to 500, while the improvement gains decrease when approaching the size of 500.
This indicates that larger embedding size are recommended, while larger embeddings can drag time efficiency and requires more computational resources, thus it is a time-efficiency trade-off.

\section{Analysis and Case Study for Symmetric/Asymmetric/Inversion Pattern}
\label{app:case_study_3}

\begin{figure}[!t]
  \centering
    \subfigure[\textit{relation:Consult}]{\includegraphics[width=0.23\textwidth]{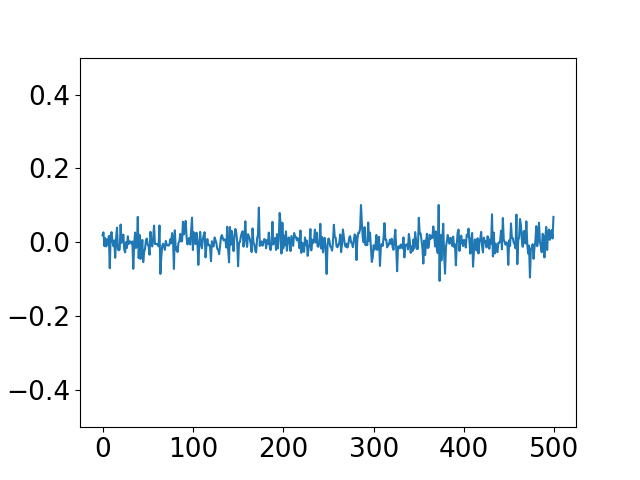}} 
	\subfigure[\textit{relation:Engage in negotiation}]{\includegraphics[width=0.23\textwidth]{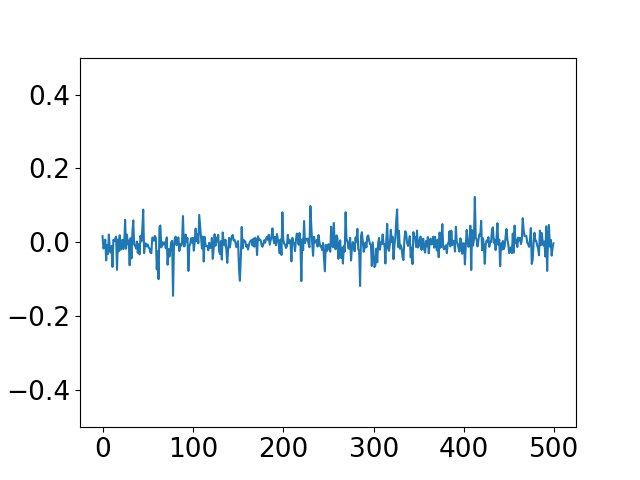}} \\
  \caption{Illustrations of the real parts ($Re$) closing to $\vect{0}$ for two symmetric relations, \emph{Consult} and \emph{Engage in negotiation}, in quaternion embeddings with size 500.}
	\label{figure:sym}
\end{figure}

\begin{figure}[!t]
  \centering
    \subfigure[\textit{relation:Threaten to halt mediation}]{\includegraphics[width=0.23\textwidth]{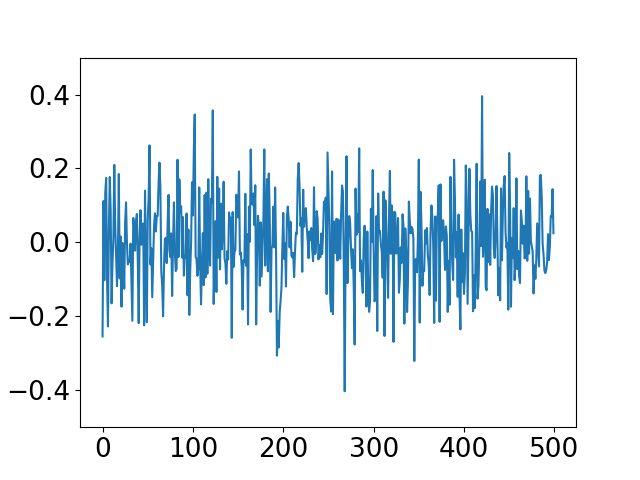}} 
	\subfigure[\textit{relation:Demand policy change}]{\includegraphics[width=0.23\textwidth]{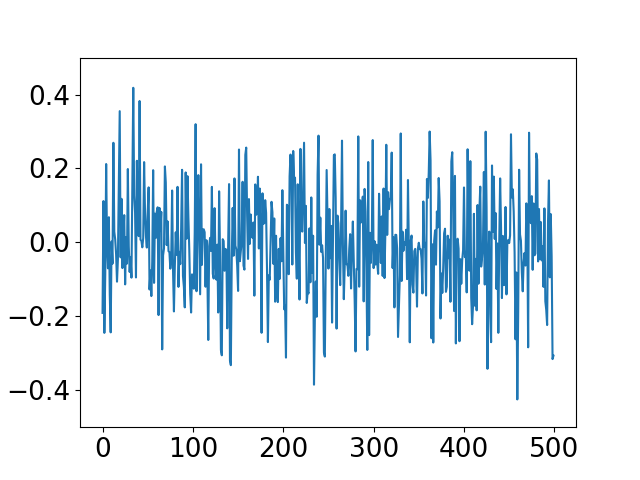}} \\
  \caption{Illustrations of the real parts $Re(\vectr)\neq\vect{0}$ for two asymmetric relations in quaternion embeddings.
}
	\label{figure:asym}
\end{figure}

\subsection{Symmetric Pattern}
As shown in Lemma \ref{lemma:symmetric} and its proof (see Appendix \ref{app:proof_symmetric}), if $r$ is a symmetric relation, we have
$\vectr + \overline{\vectr}=\vect{0}
\Leftrightarrow
    Re(\vectr)= \vect{0}$.
That is if $r$ is a symmetric relation, the real part of quaternion representation of $r$ is close to zero.
To empirically study the learned quaternion embedding of $r$, we illustrate the real parts of quaternion embeddings in Figure~\ref{figure:sym} for two symmetric relations, \textit{Consult} and \textit{Engage in negotiation}. For \exam{Consult}, we have \textit{(France, Consult, Canada, 2014-10-23)} and \textit{(Canada, Consult,  France, 2014-10-23)}.
For \exam{Engage in negotiation}, we have \textit{(Victor Ponta, Engage in negotiation, Klaus Johannis, 2014-11-11)} and \textit{(Klaus Johannis, Engage in negotiation, Victor Ponta, 2014-11-11)}.
These suggest that the relation \textit{Consult} and the relation \textit{Engage in negotiation} are two symmetric relations.
We observe that the learned quaternion embeddings in Figure~\ref{figure:sym} are close to $\vect{0}$, which confirms the ability of our model.

\begin{figure}[!t]
  \centering
    \subfigure[$Re(\vect{r_1}) + Re(\vect{r_2}) = \vect{0}$]{\includegraphics[width=0.23\textwidth]{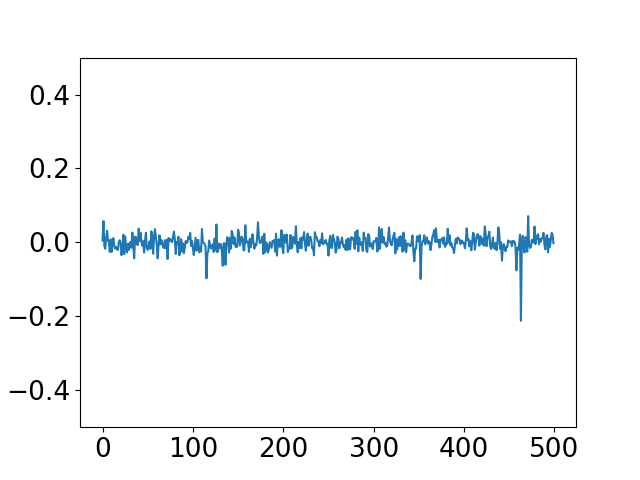}} 
	\subfigure[$\vecti(\vect{r_1}) - \vecti(\vect{r_2}) = \vect{0}$]{\includegraphics[width=0.23\textwidth]{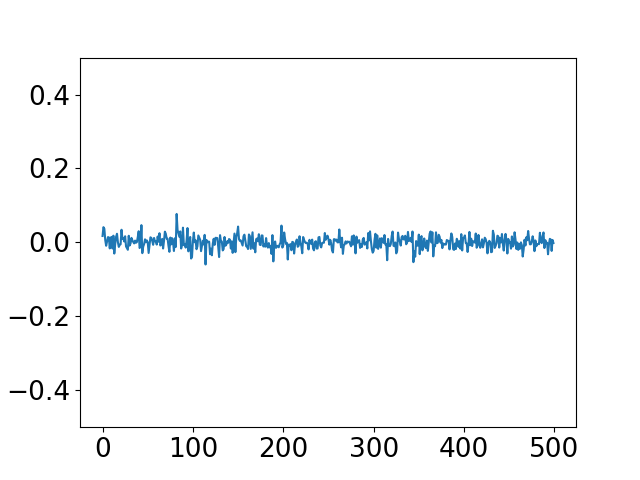}} \\
	\subfigure[$\vectj(\vect{r_1}) - \vectj(\vect{r_2}) = \vect{0}$]{\includegraphics[width=0.23\textwidth]{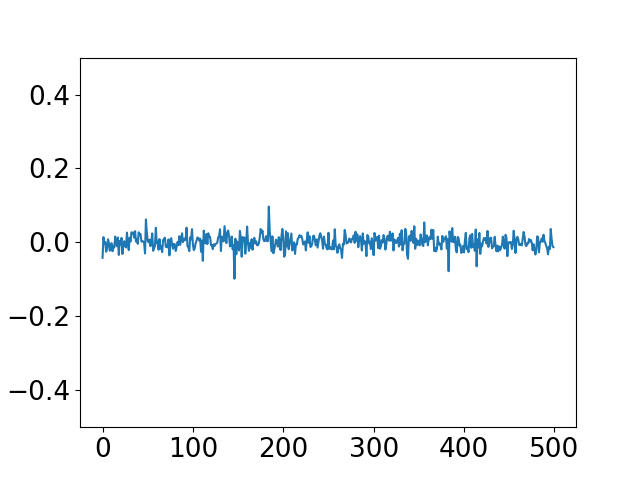}} 
	\subfigure[$\vectk(\vect{r_1}) - \vectk(\vect{r_2}) = \vect{0}$]{\includegraphics[width=0.23\textwidth]{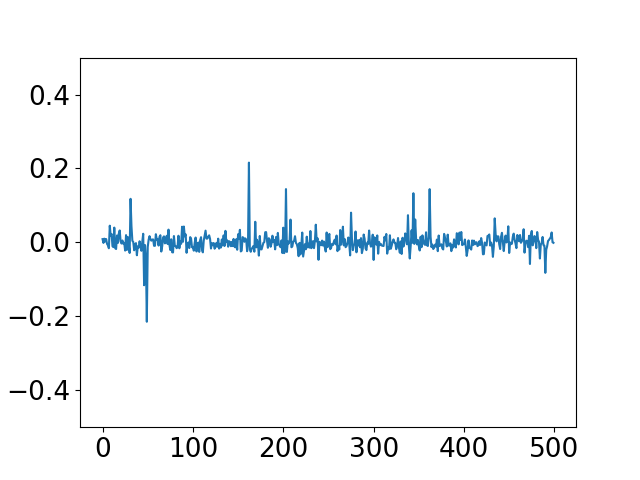}} \\
  \caption{Quaternion representations of Equation~\ref{equation:inverse}, with the real part ($Re$) and three imaginary parts ($\vecti$, $\vectj$, and $\vectk$) for an inverse relation pair: \exam{Make a visit} vs. \exam{Host a visit}.}
	\label{figure:inverse}
\end{figure}

\subsection{Asymmetric Pattern}
Opposite to symmetric pattern, if $r$ is an asymmetric relation, we have
$Re(\vectr)\neq \vect{0}$.
Intuitively, if $r$ is an asymmetric relation, the real part of quaternion representation of $r$ is supposed to be far away from zero.
Since we have \textit{(Nabih Berri, Threaten to halt mediation, Israeli Defense Forces, 2014-05-12)} and \textit{(Islamic Preacher (Iran), Demand policy change, Iran, 2014-03-02)}, the two relations \textit{Threaten to halt mediation} and \textit{Engage in negotiation} are taken as two asymmetric relations.
Figure \ref{figure:asym} illustrates the real parts of quaternion representation of them.
These observations from Figure~\ref{figure:sym} and Figure~\ref{figure:asym} show that our RotateQVS can effectively model the symmetry and asymmetry patterns and can distinguish them.

\subsection{Inversion Pattern}

Lemma \ref{lemma:inversion} and its proof (see Appendix \ref{app:proof_inversion}) show that if the relation $r_1$ is the inverse of the relation $r_2$, we have $Re(\vect{r}_1) + Re(\vect{r}_2) = \vect{0}$ and $Im(\vect{r}_1) - Im(\vect{r}_2) = \vect{0}$.
From two existing facts \textit{(Romania, Host a visit, Evangelos Venizelos, 2014-02-20)} and \textit{(Evangelos Venizelos, Make a visit, Romania, 2014-02-20)} in ICEWS14, we can find out the relation \textit{Host a visit} is the inverse of the relation \textit{Make a visit}, which satisfies the Definition \ref{definition:r4}.
Figure~\ref{figure:inverse} shows the quaternion representation as in Equation~ \ref{equation:inverse} for the inverse relation pair, containing the real part and the imaginary part (3D), where $\vecti$, $\vectj$ and $\vectk$ denote three directions of the imaginary parts.
This demonstrates our method has effective modeling for inverse relations.

\end{document}